%% file: main-paper.tex
\documentclass[conference]{IEEEtran}
\IEEEoverridecommandlockouts

\usepackage{cite}
\usepackage{amsmath,amssymb,amsfonts}
\usepackage{algorithmic}
\usepackage{graphicx}
\usepackage{textcomp}
\usepackage{xcolor}
\usepackage{comment}
\def\BibTeX{{\rm B\kern-.05em{\sc i\kern-.025em b}\kern-.08em
    T\kern-.1667em\lower.7ex\hbox{E}\kern-.125emX}}

\usepackage{booktabs}
\usepackage{colortbl}
\usepackage{footnote}
\usepackage{accents}
\usepackage{multirow}
\usepackage{makecell}
\usepackage{subfig}

\usepackage{mathrsfs} 

\newcommand{\textvec}[1]{\mathbf{#1}}

\begin{document}

\title{Continual Few-shot Adaptation for Synthetic Fingerprint Detection
}

\author{
\IEEEauthorblockN{Joseph Geo Benjamin, Anil K. Jain, and Karthik Nandakumar}
\IEEEauthorblockA{\textit{Michigan State University, East Lansing, MI, USA}}
\texttt{\footnotesize benja161@msu.edu, jain@msu.edu, nandakum@msu.edu}
}

\maketitle
\begin{abstract}
The quality and realism of synthetically generated fingerprint images have increased significantly over the past decade fueled by advancements in generative artificial intelligence (GenAI). This has exacerbated the vulnerability of fingerprint recognition systems to data injection attacks, where synthetic fingerprints are maliciously inserted during enrollment or authentication. Hence, there is an urgent need for methods to detect if a fingerprint image is real or synthetic. While it is straightforward to train deep neural network (DNN) models to classify images as real or synthetic, often such DNN models overfit the training data and fail to generalize well when applied to synthetic fingerprints generated using unseen GenAI models. In this work, we formulate synthetic fingerprint detection as a \textit{continual few-shot adaptation} problem, where the objective is to rapidly evolve a base detector to identify new types of synthetic data. To enable continual few-shot adaptation, we employ a combination of binary cross-entropy and supervised contrastive (applied to the feature representation) losses and replay a few samples from previously known styles during fine-tuning to mitigate catastrophic forgetting. Experiments based on several DNN backbones (as feature extractors) and a variety of real and synthetic fingerprint datasets indicate that the proposed approach achieves a good trade-off between fast adaptation for detecting unseen synthetic styles and forgetting of known styles. 


\end{abstract}


\section{Introduction}
\label{sec:intro}
Fingerprint recognition systems (FRS) are widely used in many applications including law enforcement, border control, welfare disbursement, and access control for personal devices. It is well-known that the integrity of FRS can be compromised in several ways \cite{intro-mansfield2002best}. Specific threats such as presentation attacks \cite{pad-chugh2018fingerprint} and template leakage \cite{intro-jain2005biometric} have been studied in detail, and techniques to counteract such threats have been proposed. In this work, we consider the \textit{vulnerability of FRS to data injection attacks} (ISO/IEC 25456\cite{ISOIEC25456}, CEN/TS 18099\cite{CENTS18099}), where synthetic images are inserted into the FRS pipeline before the feature extraction process. Such digital injection attacks can severely compromise the integrity of FRS and lead to false leads/wrongful arrests in law enforcement applications or unauthorized access in civilian applications. Though data injection not a new threat, \textit{its severity has increased dramatically in recent years due to rapid advancements in generative artificial intelligence (GenAI)}. 

\begin{figure}
    \centering
    \includegraphics[width=\linewidth]{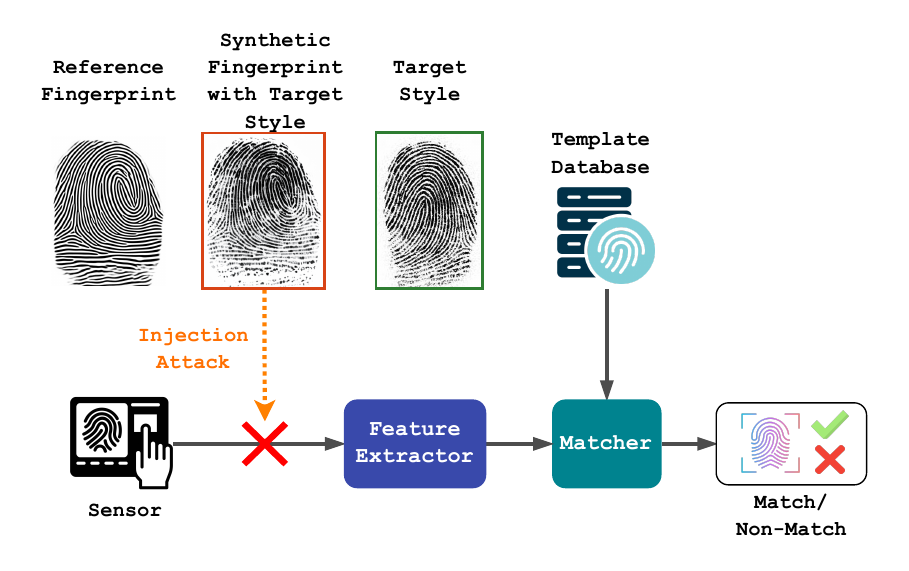}
    \caption{Vulnerability of fingerprint recognition systems (FRS) to data injection attacks. With advancements in GenAI, it is now possible to generate synthetic fingerprint images with any target style, which can be used to fool any FRS (especially in remote authentication scenarios) via bypassing of the sensor.}
    \label{fig:intro-image}
\end{figure}

Methods for the generation of synthetic fingerprints have been around for more than two decades \cite{sfinge-cappelli2002synthetic} and these synthetic fingerprint images have been primarily used to develop better fingerprint recognition algorithms \cite{printsgan-engelsma2022printsgan, iwgan-mistry2020fingerprint}. However, the quality of synthetic fingerprint images created using traditional generation methods\cite{sfinge-cappelli2002synthetic} was often low and the generation process lacked precise control. Hence, it was easier to identify such synthetic fingerprint images through visual inspection or simple automated filters. Recent GenAI models can create synthetic fingerprint images with a high degree of realism and precision \cite{fpgan-shoshan2024fpgan}. For example, the diffusion-based GenPrint framework \cite{genprint-grosz2024universal} can take any reference fingerprint image and regenerate it in real-time so that the output closely resembles a fingerprint image acquired using a specific fingerprint sensor under specific acquisition conditions (see Figure \ref{fig:intro-image}). Although data injection attacks can be potentially mitigated by ensuring end-to-end security of the entire pipeline (from sensor to the final match decision), this is not always possible, especially in remote authentication scenarios. Thus, the ability to accurately detect and flag such synthetic fingerprints becomes a critical element of FRS security.

While synthetic fingerprint detection is an unexplored problem, it has many parallels to fingerprint presentation attack detection (PAD) \cite{pad-rattani2015open}, AI-generated image detection \cite{aigen-wufew}, and deepfake image/video/audio detection \cite{deepfakes-cao2022end}. Similar to these related problem domains, it is relatively straightforward to learn a deep neural network (DNN) model to classify the given fingerprint image as real or synthetic. However, this approach mainly relies on the assumption that the attack/generated images contain some common artifacts that are not present in real images and the DNN model can reliably learn to detect these artifacts from the training datasets. The hope is that the detector can achieve universal \textit{zero-shot generalization} against unseen attacks/generation models. But this is seldom the case in practice because of the continuous evolution of GenAI models and fingerprint sensors, which leads to large domain shifts that cannot be handled with static representations and decision boundaries learned based on historical data. Consequently, there is typically a catastrophic collapse in detection accuracy when encountering synthetic images created using unseen generators or real images acquired using novel sensors.

In this work, we propose a DNN-based synthetic fingerprint detector coupled with an efficient few-shot adaptation strategy to enable detection of synthetic images with novel styles. The contributions of this work are three-fold: 
\begin{itemize}
    \item To the best of our knowledge, this is \textit{the first work} to explore the problem of synthetic fingerprint detection. We formulate the problem as a continual few-shot adaptation task, aiming to continuously evolve a base detector to recognize synthetic fingerprints with unseen styles.
    \item  To facilitate smoother evolution of the synthetic fingerprint detector, we use a combination of binary cross-entropy and supervised contrastive losses, and replay a small set of samples from previously encountered styles during fine-tuning to reduce catastrophic forgetting.
    \item Experiments conducted using multiple DNN backbones and a diverse collection of real and synthetic fingerprint datasets show that the proposed method effectively balances rapid adaptation to unseen synthetic styles while retaining knowledge of previously learned styles.
\end{itemize}

\section{Background}
\subsection{Related Works}
\noindent\textbf{\textit{Synthetic Fingerprint Generation:}} Synthetic fingerprints help in addressing the scarcity of large-scale fingerprint datasets needed for developing fingerprint recognition systems. Many rule-based, statistical, handcrafted methods have been proposed for synthetic fingerprint generation, such as SFinGe \cite{sfinge-cappelli2002synthetic}. 
With the rise of modern deep learning–based generative models, generative adversarial networks (GANs) have become a dominant approach for synthetic fingerprint generation \cite{printsgan-engelsma2022printsgan, fpgan-shoshan2024fpgan, iwgan-mistry2020fingerprint, gan-shukla2024vikriti}. GANs learn to model complex data distributions directly from real samples, producing better realism and richer textural details. More recently, diffusion models have emerged as a powerful alternative for the image generation process. Diffusion-based generation\cite{genprint-grosz2024universal} can produce highly realistic fingerprints that emulate specific acquisition conditions while maintaining strong structural consistency. The growing realism of synthetic data poses security risks as adversaries can insert hard-to-detect fabricated images into biometric databases.

\noindent\textbf{\textit{Fingerprint Presentation Attack Detection:}} Fingerprint PAD is a heavily researched topic where spoofing happens using physically fabricated fingerprints or artificial materials to impersonate users\cite{PAD-marcel2019handbook}.
Early works\cite{pad-rattani2015open} used texture and minutiae-based descriptors for spoof detection. Later, CNNs achieved state-of-the-art performance by learning discriminative features directly from images \cite{pad-chugh2018fingerprint}. Generalization across fingerprint sensors remains a key challenge in PAD. \cite{pad-chugh2019generalization} tackles this by generating variations of spoof materials using a generative model, while later work \cite{pad-grosz2020fingerprint} employs adversarial training to learn domain-invariant representations and mitigate sensor variability.
PAD approaches often rely on texture artifacts induced by fabrication, which become less effective as generative models produce more realistic fingerprints.

\noindent\textbf{\textit{Continual Learning (CL):}} 
CL aims to continually update DNNs to a sequence of new tasks while preventing catastrophic forgetting. Initial works focused on regularization-based approaches\cite{cl-ewc_kirkpatrick2017overcoming}. Replay-based approaches also gained traction by maintaining a small episodic memory of past samples, which are replayed during training to mitigate catastrophic forgetting \cite{cl-replay_chaudhry2019tiny_episodes}. Other representation-centric methods focus on learning task-invariant features or subspace constraints, improving stability\cite{cl-hou2019learning_represent}. Then, architecture-based methods dynamically expand networks per task to avoid forgetting\cite{cl-yoon2017lifelong_expand}. We treat synthetic images generated by newer generative models as domain-shifted distributions of previously seen synthetic data. Hence, our work focuses on the domain-incremental learning scenario.

\noindent\textbf{\textit{FewShot Learning (FSL):}} 
FSL aims to enable models to learn new tasks from just a few labeled examples, allowing them to quickly adapt to unseen tasks in practice. Metric-learning methods approaches such as Prototypical Networks\cite{fewshot-snell2017prototypical} and Matching Networks\cite{fewshot-vinyals2016matching} learn embedding spaces that enable fast adaptation. Optimization-based approaches like MAML\cite{maml-finn2017model} learns a meta-initialization for rapid adaptation to new tasks with few examples, which was further simplified by ANIL\cite{anil-raghu2019rapid} by updating only the task-specific head.

\noindent\textbf{\textit{DeepFakes and AI-Generated Images Detection:}}
Deepfake detection has become an increasingly important research field as morphed content with increasing realism undermines authenticity and trust in society. Early approaches \cite{deepfake-rossler2019faceforensics++} focused more on identifying subtle visual artifacts using CNNs, whereas more recent methods \cite{deepfakes-cao2022end} leverage reconstruction-based learning combined with classification to detect fake images. More broadly, detecting AI-generated or manipulated images extends the scope of research beyond deepfakes to cover diverse forms of synthetic media. Many works focus on detecting artifacts and spectral patterns arising from the generation process \cite{aigen-karageorgiou2025any}. However, the rapid evolution of generative models introduces an additional challenge, thus recent studies \cite{aigen-wufew} address the challenge by learning the newer generative distributions in a few-shot adaptation paradigm.

\section{Methodology}

\subsection{Preliminaries}
\label{sec:prelims}

\noindent \textbf{Notations}: Let $\mathcal{X}$ denote the super-set of all fingerprint images. Let $\mathcal{R} \subset \mathcal{X}$ indicate the subset of real fingerprint images. A real fingerprint image is obtained from an individual using an acquisition process (e.g., rolled, plain, latent, etc.) under certain acquisition conditions (e.g., ink on paper, sensor used for livescan capture, latent surface/characterization method, etc.). Similarly, let $\mathcal{S} \subset \mathcal{X}$ denote the subset of synthetic fingerprint images. A synthetic fingerprint image is generated using a generative model (e.g., \emph{GAN} \cite{fpgan-shoshan2024fpgan, printsgan-engelsma2022printsgan, iwgan-mistry2020fingerprint}, diffusion model \cite{genprint-grosz2024universal}, or other hand-crafted methods \cite{sfinge-cappelli2002synthetic, ibg_unrel-cao2018fingerprint}) with some generation conditions (fingerprint attributes specified to the generator, often via a text prompt). The combination of acquisition process and acquisition condition for real images, or the generative model and generation condition for synthetic images is referred to as the \textit{style} of the fingerprint image.


\noindent \textbf{Problem Statement}: Suppose that we are given an initial training set of $N$ fingerprint images $\mathcal{D}_0 = \{(\mathbf{x}_i, y_i,\mathbf{c}_i)\}_{i=1}^{N}$, where $y_i = 0$ indicates a real image, $y_i = 1$ indicates a synthetic image, and $\mathbf{c}_i$ denotes the image style. Let $\mathbf{C}_0 = \cup_{i=1}^{N} \mathbf{c}_i$ be the set of all real and synthetic image styles included in the training set. The overarching objective of this work is to design a synthetic fingerprint detector $\mathcal{M}_{\theta}:\mathcal{X} \rightarrow \{0,1\}$ that determines if the given test fingerprint image $\mathbf{x} \in \mathcal{X}$ is real (i.e., $\mathbf{x} \in \mathcal{R}$) or synthetic (i.e., $\mathbf{x} \in \mathcal{S}$). In practice, the detector $\mathcal{M}_{\theta}$ often produces a real-valued score $s \in [0,1]$ and $\mathcal{M}_{\theta}(\mathbf{x}) = I(s > \tau)$, where $I$ is an indicator function (outputs $1$ if the condition is true and $0$ otherwise) and $\tau \in [0,1]$ is the decision threshold. Such a detector can make two types of errors: \textit{false detection} (real image categorized as synthetic) and \textit{missed detection} (failure to detect a synthetic image), which can be characterized using false detection rate (FDR) and true detection rate (TDR) metrics, respectively. Thus, 

\begin{equation}
\begin{aligned}
    FDR(\tau) :=  \mathbb{P}\big[s \geq \tau |\; \mathbf{x} \in \mathcal{R}\big] \\
    TDR(\tau) :=  \mathbb{P}\big[s \geq \tau |\; \mathbf{x} \in \mathcal{S}\big]
\end{aligned}
\end{equation}

\noindent where $\mathbb{P}$ denotes probability. The primary goal is to design the detector $\mathcal{M}_{\theta}$ such that it \textit{maximizes TDR while minimizing FDR}. Ideally, the synthetic fingerprint detector must be agnostic to the image style. More specifically, let $\mathbf{c}$ be the style of the given test image $\mathbf{x}$. If $\mathbf{c} \in \mathbf{C}_0$, then the test image is considered to be a \textit{known style}. Otherwise, it is considered to be a \textit{novel/unseen} style. The detector can be considered as \textit{generalizable} if it achieves \textit{a high TDR at a given FDR for samples with unseen styles}. However, our preliminary analysis indicates that such zero-shot generalization is hard to achieve in the case of synthetic fingerprint detection due to the wide variability among generative models. Furthermore, more advanced synthetic fingerprint generators continue to be developed. Hence, we formulate the problem as\textbf{ \textit{continual few-shot adaptation}}. Let $\mathcal{M}_{\theta_0}$ be the base detector learned using the initial training set $\mathcal{D}_0$. Given a small number of samples from an unseen style, our goal is to dynamically adapt the detector such that it can perform well for both the new style and all previously known styles. 

\subsection{Proposed Method}

\begin{figure}
    \centering
    \includegraphics[width=\linewidth]{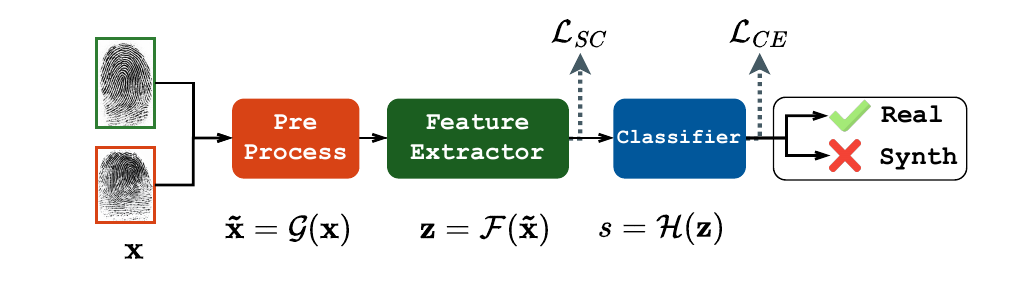}
    \caption{Illustration of the proposed approach for synthetic fingerprint detection.}
    \label{fig:method}
\end{figure}

We propose employing a DNN-based framework for synthetic fingerprint detection that has three main components as shown in Figure \ref{fig:method}. First, a pre-processing function $\mathcal{G}$ is applied to map the input fingerprint image $\mathbf{x}$ of arbitrary size to obtain an RGB image of fixed size containing $H \times W$ pixels. In this work, the pre-processor $\mathcal{G}$ detects the foreground fingerprint region in the given image and crops a $H \times W$ window that is mostly contained within the foreground region. Let $\textvec{\widehat{x}} \in \mathbb{R}^{H \times W \times 3}$ denote the pre-processed image obtained from $\mathbf{x}$. A pre-trained DNN-based backbone is used as the feature extractor $\mathcal{F}_{\phi}: \mathbb{R}^{H \times W \times 3} \rightarrow \mathbb{R}^d$ to extract a fixed-length representation $\mathbf{z} \in \mathbb{R}^d$ from $\textvec{\widehat{x}}$, where $d$ denotes the dimensionality of the feature vector. A classifier head $\mathcal{H}_\psi:  \mathbb{R}^d \rightarrow [0,1]$ is then used to map the feature representation to a real-valued score $s \in [0,1]$, which represents the probability of input image being synthetic. Thus, $s = \mathcal{M}_{\theta}(\mathbf{x}) = \mathcal{H}_{\psi}(\mathcal{F}_\phi(\mathcal{G}(\mathbf{x})))$ and the set of learnable parameters are $\theta := \{\phi,\psi\}$.

\input{tables/datasets}

\noindent \textbf{Learning the base detector}: Given the initial training set of $\mathcal{D}_0$, the parameters of feature extractor and classifier head can be learned by minimizing the binary cross-entropy (CE) loss.

\begin{equation}
    \mathcal{L}_{CE} = -\frac{1}{N}\sum_{i=1}^{N} [y_i \log{(s_i)} + (1-y_i) \log{(1-s_i)}].
\end{equation}

\noindent Thus, the optimization problem for learning the base detector can be summarized as:

\begin{equation}
\label{eq:opt}
    \theta_0 \leftarrow {\arg\min}_{\theta} ~ \mathcal{L}_{CE} (\mathcal{D}_0).
\end{equation}

\noindent During the above optimization, $\phi$ is initialized based on a pre-trained model and $\psi$ is initialized randomly.

\noindent \textbf{Continual few-shot adaptation of the detector}: Suppose that we are presented with a sequence of $K$ datasets, $\mathcal{D}_1, \mathcal{D}_2, \cdots \mathcal{D}_K$, where each dataset $\mathcal{D}_k$ contains a few samples $N_k$ belonging to an unseen style $\mathbf{c}_k$, i.e., $\mathbf{c}_k \notin (\mathbf{C}_0 \cup_{\tilde{k} =1}^{k-1} \mathbf{c}_{\tilde{k}})$. Note that these new datasets could correspond to real images from a new sensor or synthetic images from an unseen generator. Let $\mathcal{D}_k = \{(\mathbf{x}_{jk}, y_{jk} ,\mathbf{c}_{jk})\}_{j=1}^{N_k}$, where $y_{jk} = 0/1$ and $\mathbf{c}_{jk} = \mathbf{c}_{k}$, $\forall ~ j = 1,\cdots,N_k$ and $k = 1,\cdots,K$. While it is possible to directly adapt the current detector parameters $\theta_{k-1}$ using the set $\mathcal{D}_k$ by applying Eq. (\ref{eq:opt}), this approach leads to the well-known \textit{catastrophic forgetting} problem \cite{cl-ewc_kirkpatrick2017overcoming}, where the model overfits the new data and forgets the decision boundaries learned for the previously known styles. 

To mitigate forgetting, we introduce two changes to facilitate smoother adaptation. First, we apply the experience replay framework \cite{cl-replay_chaudhry2019tiny_episodes}, where we retain a few samples for each of the previously known styles in the form of a memory buffer and add them to the new samples to form $\mathcal{\tilde{D}}_{k} = (\mathcal{\tilde{D}}_{0} \cup_{\tilde{k} =1}^{k} \mathcal{D}_{\tilde{k}})$, where $\mathcal{\tilde{D}}_{0} \subset \mathcal{D}_{0}$. Though the number of samples in $\mathcal{\tilde{D}}_{0}$ is much smaller than $N$, care is taken to ensure that a minimum number of samples ($N_0$) from all available styles in the initial training dataset are included in $\mathcal{\tilde{D}}_{0}$. Let $\tilde{N}$ be the total number of samples in $\mathcal{\tilde{D}}_{k}$. To further aid the fine-tuning process, we apply the supervised contrastive (SC) loss \cite{supcon-khosla2020supervised}, which pulls the representations of the same class together and simultaneously pushes the representations of different classes far apart.

\begin{equation}\label{eq:supcon}
    \mathcal{L}_{SC} = \sum_{i=1}^{\tilde{N}}\frac{1}{(\tilde{N}-1)}\sum_{\substack{j=1 \\ j \neq i, y_j = y_i}}^{\tilde{N}}\log\left(\frac{\exp{(\mathbf{z}_i \cdot \mathbf{z}_j/\beta)}}{\sum_{\substack{\ell=1 \\ \ell \neq i}}^{\tilde{N}}\exp{(\mathbf{z}_i \cdot \mathbf{z}_\ell/\beta)}}\right),
\end{equation}

\noindent where $\beta$ is the temperature parameter. Thus, the optimization problem for updating the detector can be summarized as:

\begin{equation}
    \theta_k \leftarrow {\arg\min}_{\theta} ~ \mathcal{L}_{CE} (\mathcal{\tilde{D}}_k) + \lambda \mathcal{L}_{SC} (\mathcal{\tilde{D}}_k),
\end{equation}

\noindent where the hyperparameter $\lambda$ balances the two losses. Note that the previous version of the model parameters $\theta_{k-1}$ is used as the initialization for the above optimization.

\section{Experiments}
\input{tables/results-table}

\subsection{Datasets Used}
For real fingerprint data, we utilize impressions collected from multiple datasets that capture variations in acquisition conditions and sensor characteristics. In this study, we assume that the underlying distribution of real impressions is fixed and no novel sensor is introduced during evaluation. Hence, we pool all real datasets listed in Table~\ref{tab:fingerprint-datasets} and randomly split the data into train-test partitions, reserving 10\% from each dataset for testing and using the remaining 90\% for initial training of base detector. We retain $N_0 = 100$ randomly sampled images per dataset for replay during continual few-shot adaptation.

For synthetic data, we use six datasets summarized in Table~\ref{tab:fingerprint-datasets}, covering a wide range of generation methods. Owing to its substantial diversity in impression types and sensor styles, 90\% of the samples generated using GenPrint~\cite{genprint-grosz2024universal} are used to train the base detector and the remaining 10\% of this data is reserved for evaluation. For the remaining synthetic datasets, we use only $N_k = 100$ samples for continual adaptation, while all remaining samples are held out as the test set.

\subsection{Implementation Details}

For all main experiments and ablation studies (except Fig.~\ref{fig:backbone-perform}), we use a ViT-Small model initialized with ImageNet pretraining as the feature extractor $\mathcal{F}$. The class token output by the final block of the ViT is considered as the feature representation ($d = 384$). A multi-layer perceptron (MLP) is used as the classification head $\mathcal{H}$. Specifically, $\mathcal{H}$ has two fully connected layers: hidden layer has $(d/2)$ nodes with ReLU activation and there is one output node. AdamW is used as the optimizer with a learning rate of $10^{-5}$, weight decay of 0.01, and a cosine annealing schedule with a minimum learning rate of $10^{-6}$. All other optimization parameters are kept at PyTorch default values. We train the base detector for 5 epochs with a batch size of 256 and select the best-performing model across epochs. During continual few-shot adaptation, the batch size is reduced to 64  and fine-tuning is performed for 10 epochs. During adaptation, we set $\lambda = 0.1$ and $\beta = 0.1$.

\subsection{Training and Adaptation Protocol}
As indicated earlier, the base detector training is performed based on 90\% of samples from all real datasets and synthetic images generated using GenPrint. After this initial training, novel synthetic datasets are introduced in the following order for continual adaptation: SFinGe\cite{sfinge-cappelli2002synthetic}, IBG-Novetta\cite{ibg_unrel-cao2018fingerprint}, IWGAN\cite{iwgan-mistry2020fingerprint}, PrintsGAN\cite{printsgan-engelsma2022printsgan}, and FPGAN\cite{fpgan-shoshan2024fpgan}. Note that this specific choice of adaptation sequence is based on the chronological order in which these synthetic fingerprint generation methods were introduced in the literature. Consequently, one could expect the adaptation to become progressively more difficult as the latter methods are likely to generate more realistic synthetic fingerprints that are more difficult to detect. However, the test set is fixed across all the main experiments, and two main metrics are analyzed: TDR and FDR at a fixed threshold ($\tau = 0.5)$ and TDR@0.1\% FDR. 

\subsection{Results}
\noindent \textbf{Continual Few-shot Adaptation}: In this experiment, we compare the generalization capabilities of two models: ViT-CE and ViT-CE+SC. While ViT-CE represents the ViT-Small feature extractor trained and adapted using only the CE loss, ViT-CE+SC denotes training using only CE loss and adaptation using both CE and SC losses. In both cases, experience replay is employed. The objectives of this experiment are three-fold: (i) evaluate the zero-shot generalization capability of the base detector to new synthetic styles and establish the benefits of adaptation, (ii) quantify the improvement in generalization performance due to the SC loss, and (iii) measure the ability of the adapted models to retain knowledge of the previous styles. Table \ref{tab:cross_gen_results} summarizes the results of this experiment.

From Table \ref{tab:cross_gen_results} we observe that the base detector has very poor generalization to unseen synthetic styles. For examples, the base detector has $96.68\%$ TDR@$0.1\%$ FDR on the GenPrint test set, which is a known style because samples from this dataset are used in initial training. However, the TDR collapses to almost $0\%$ when the same base detector is applied to synthetic test sets with novel styles as shown in the remaining columns of the first row. This catastrophic collapse demonstrates that zero-shot generalization to unseen synthetic styles is hard to achieve in synthetic fingerprint detection.  

Next, we observe that few-shot adaptation is beneficial in improving the detection performance on new styles. As shown in the diagonal cells (highlighted in \textcolor{orange}{orange}), after each adaptation the TDR on the test set of the recently seen style improves significantly. Furthermore, we also observe a declining trend in this performance improvement. While styles with lower quality images generated by hand-crafted methods can be easily detected after adaptation, images generated using the more sophisticated GAN methods are harder to detect even after few-shot adaptation. In this scenario, continual few-shot adaptation based on a combination of CE and SC losses is certainly beneficial, because the ViT-CE+SC model consistently outperforms the ViT+CE model across all adaptation steps. This clearly demonstrates the benefits of introducing the SC loss during adaptation, which stabilizes representation learning even though the number of available samples is small. Moreover, we also observe some forward knowledge transfer, especially when there is some similarity between the synthetic styles. For example, adaptation based on SFinge improves the ability to detect IBG-Novetta style, even though the latter is completely unseen. Similarly, adaptation to IWGAN marginally increases the detection performance on the other unseen GAN styles. 

\begin{figure}[h]
    \centering
    \subfloat[Without replay, the model encounters catastrophic forgetting on previous styles after adaptation to a new style.]{
    \includegraphics[width=0.38\linewidth]{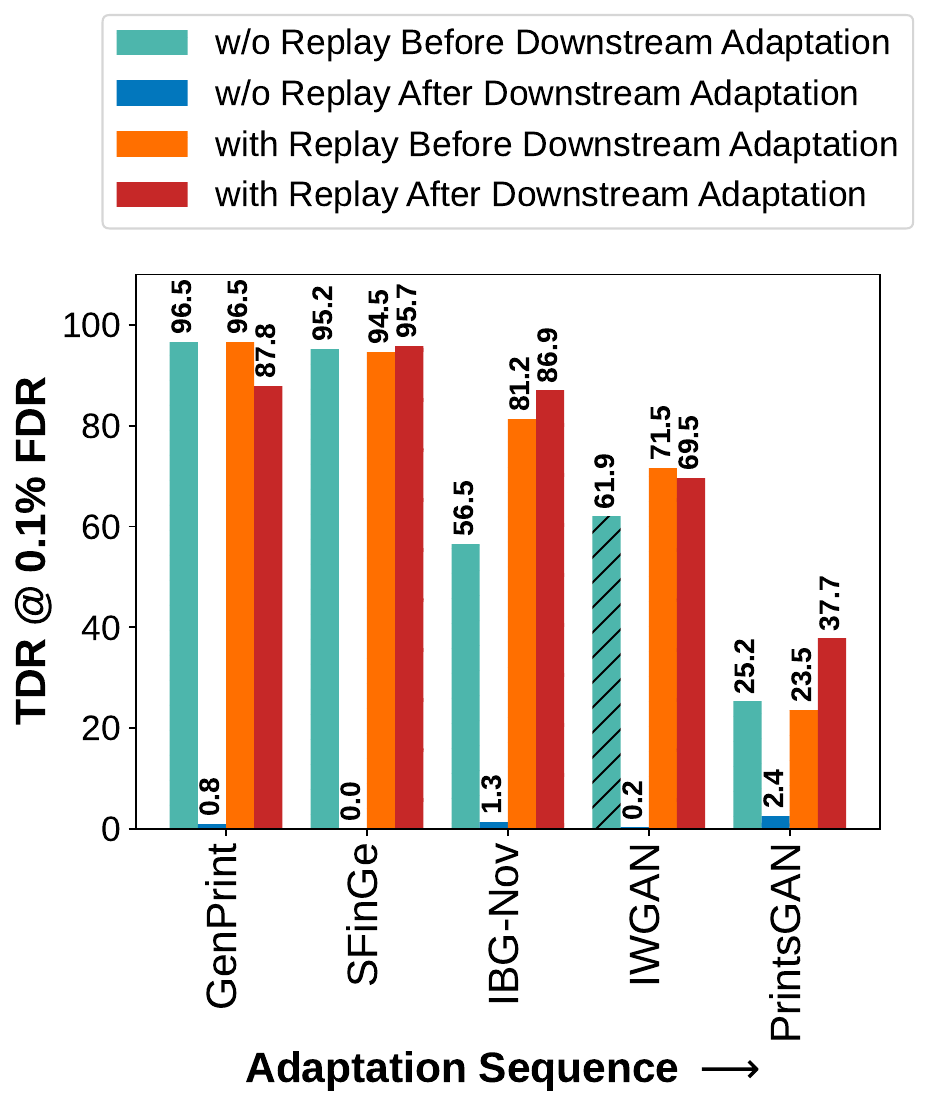}
        \label{fig:replay-effect}
    }
    \hfill
    \subfloat[Increasing sample size consistently improves the detection accuracy across all adaptation steps, especially from 10 to 50.]{
    \includegraphics[width=0.4\linewidth]{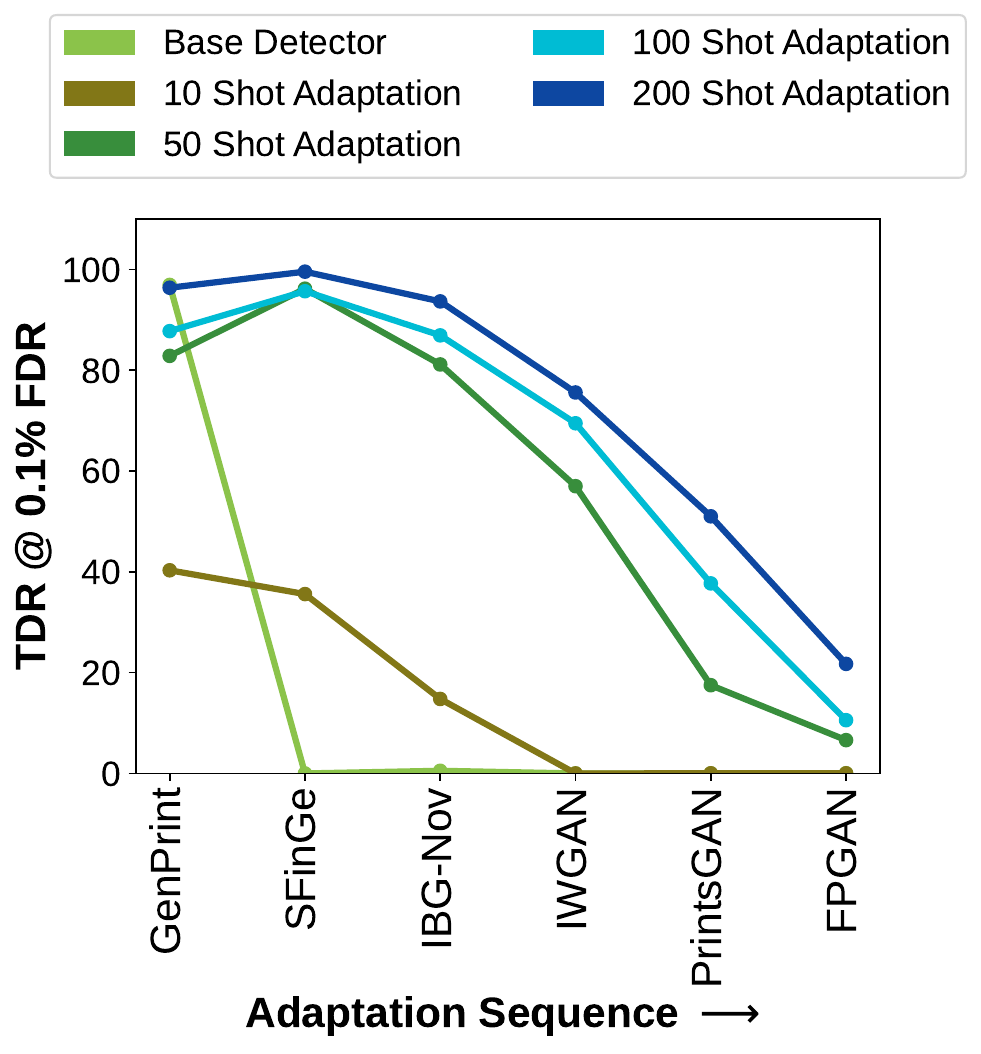}
        \label{fig:fewshot-effect}
    }
    \caption{Impact of replay and sample size on continual few-shot adaptation of the base detector.}
    \label{fig:fewshot+replay}
\end{figure}

\begin{figure}[h]
    \centering
    \subfloat[Employing more sophisticated DNN backbones for feature extraction can improve detection accuracy at higher computational cost.]{
    \includegraphics[width=0.38\linewidth]{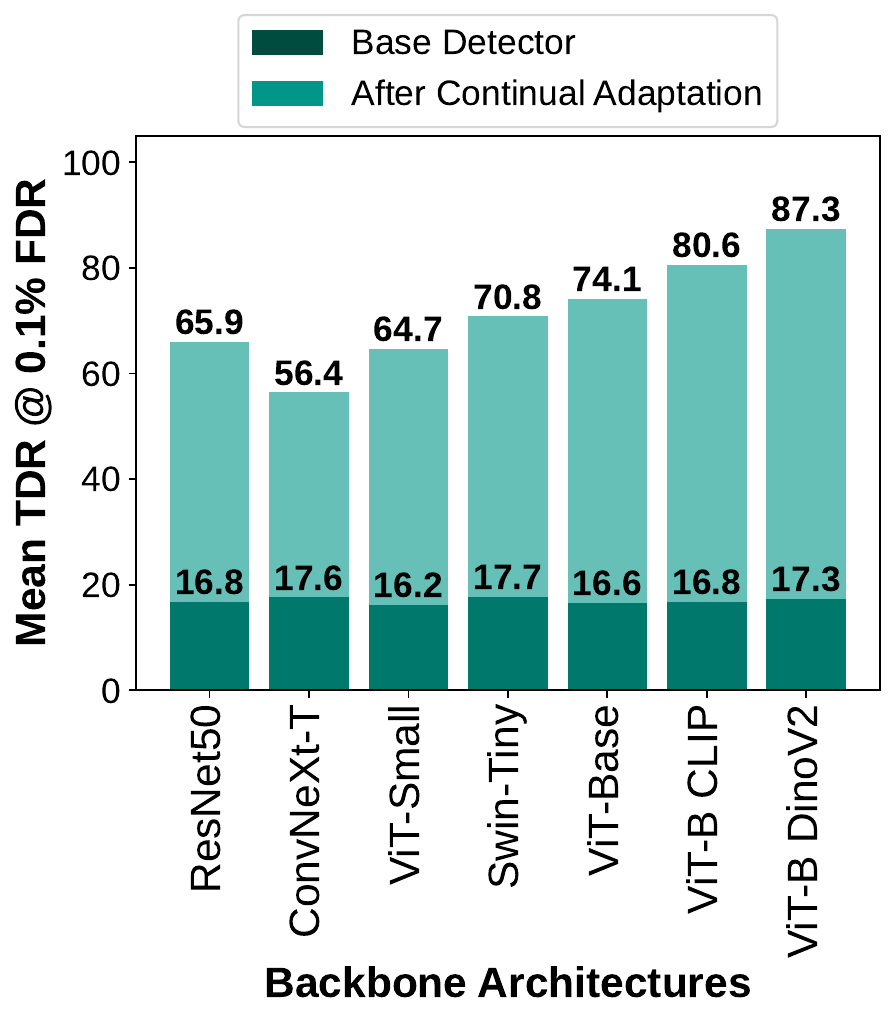}
        \label{fig:backbone-perform}
    }
    \hfill
    \subfloat[Though the value of $\lambda$ has mimimal impact on detection accuracy, $\lambda = 0.1$ achieves the best trade-off between fast adaptation and forgetting.]{
    \includegraphics[width=0.38\linewidth]{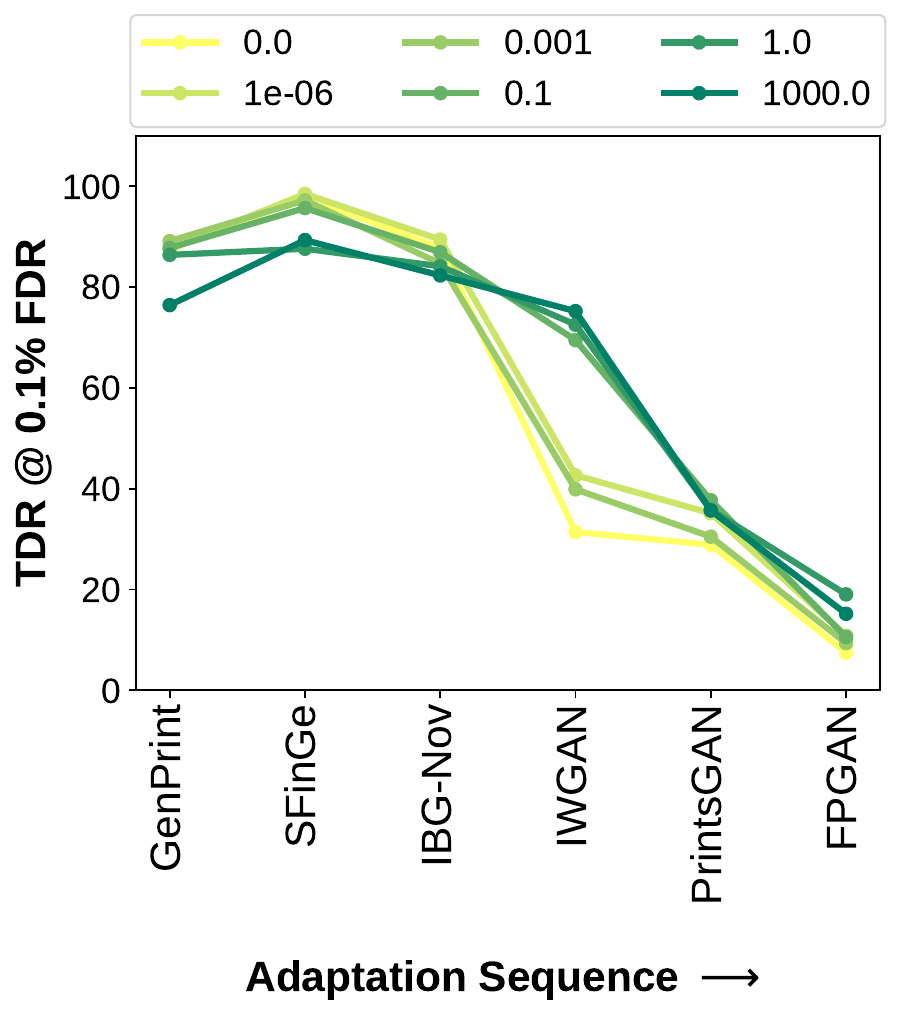}
        \label{fig:lambda-effect}
    }
    \caption{Generalization to other DNN-backbones and sensitivity to weight ($\lambda$) assigned to the supervised contrastive (SC) loss.}
    \label{fig:backbone+lambda}
\end{figure}

Finally, we observe that incorporating experience replay minimizes catastrophic forgetting to a great extent. While there is some performance degradation for some previously known styles (e.g., TDR@$0.1\%$ FDR for GenPrint decreases after multiple adaptation steps), the performance drop is not catastrophic. Interestingly, there is also some backward transfer of knowledge. For example, adaptation to the GAN models also increases the detection accuracy for the previously seen hand-crafted styles. Similarly, adaptation to FPGAN increases the performance on the previously seen PrintsGAN. However, this backward knowledge transfer is not uniform (e.g., degradation for GenPrint and IWGAN) and this can be attributed to varying similarity between style distributions.

\noindent \textbf{Impact of replay}: Our experiments show that replay plays a critical role in retaining the knowledge about previously known styles. As shown in Figure \ref{fig:replay-effect}, in the absence of replay, the performance on previous styles drops to almost zero as soon as the model is adapted to the new style. However, incorporating replay preserves performance on earlier styles to a large extent, with some cases experiencing performance improvement due to backward knowledge transfer.

\noindent \textbf{Impact of sample size in few-shot adaptation}: Increasing the number of samples available for few-shot adaptation increases the performance on the adaptation task. As observed from Figure \ref{fig:fewshot-effect}, the performance improvement is especially significant when moving from 10 to 50 samples. Though there is a clear improvement when the sample size is further increased to 100 and 200, the gains become increasingly marginal.

\noindent \textbf{Generalization to other backbones}: The proposed learning and adaptation strategy is agnostic to any DNN backbone used as the feature extractor. Results in Figure \ref{fig:backbone-perform} demonstrate that the same strategy can be applied to multiple backbones. In fact, more sophisticated backbones (e.g., ViT-Base) pre-trained using more advanced methods (e.g., CLIP \cite{clip_radford2021learning} and DINOv2 \cite{dinov2-oquab2023}) can lead to significant improvement in the detection accuracy, albeit at a higher computational cost.

\section{Conclusion}

In this work, we presented the first study on synthetic fingerprint detection and framed it as a continual few-shot adaptation problem. We have combined binary cross-entropy and supervised contrastive losses while replaying a small set of previously encountered samples to mitigate catastrophic forgetting. Extensive experiments across diverse real and synthetic fingerprint datasets demonstrate that our approach effectively balances adaptation to novel synthetic styles while improving retention of previously learned knowledge, highlighting its potential as a defense against data injection attacks.
\vspace{1pt}

\noindent \textbf{Acknowledgment}: This material is partly based on work supported by the Office
of Naval Research N00014-24-1-2168.

\bibliographystyle{IEEEtran}
\bibliography{IEEEabrv,IEEEfull}

\appendices
\onecolumn

\section{Additional Results}

\subsection{\textbf{ROC Curves:}}
\begin{figure}[!h]
    \centering
    \includegraphics[width=\linewidth]{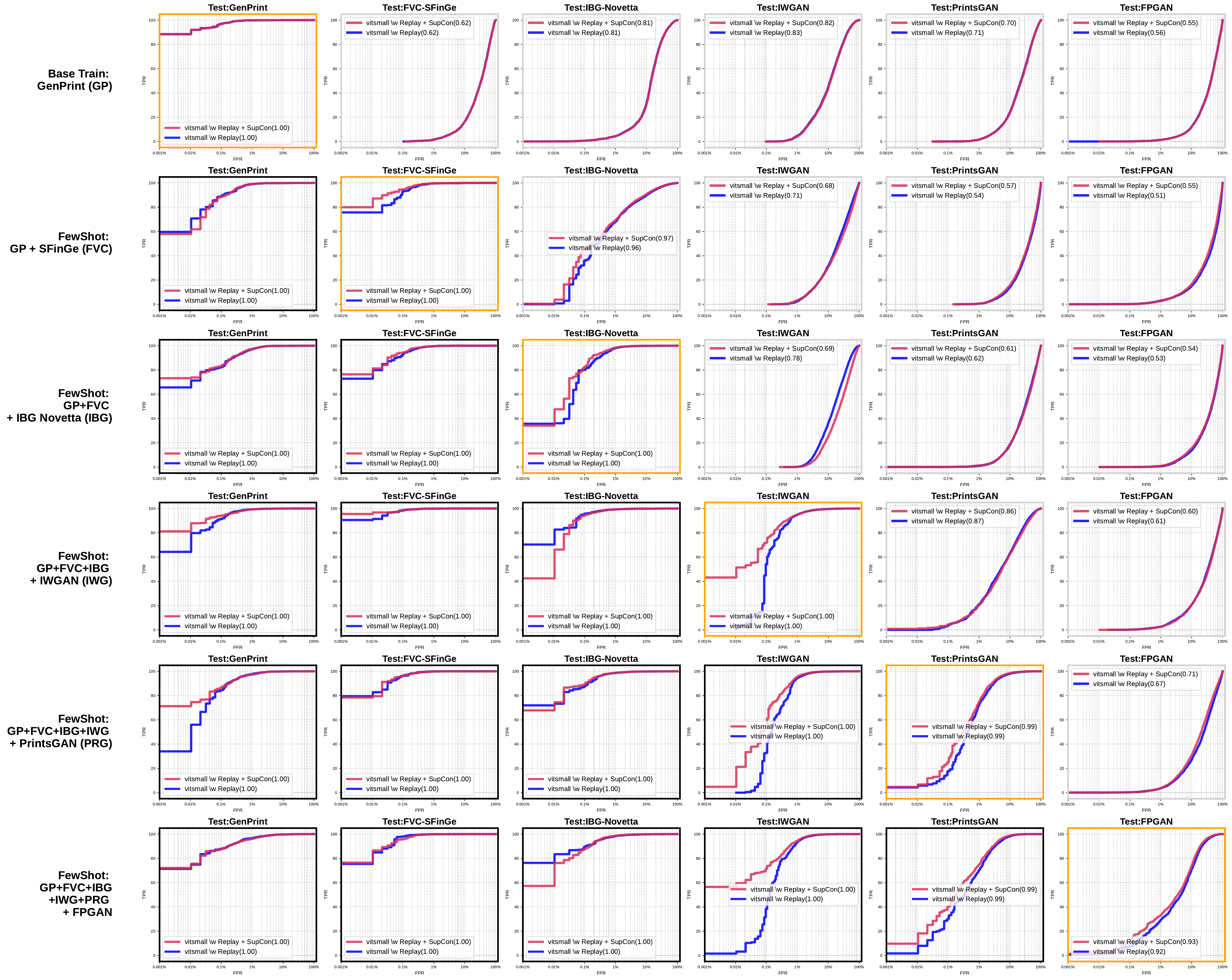}
    \caption{ROC curves (in log scale) illustrating the performance of continual few-shot adaptation. Each row corresponds to a different test set, while each column represents the incremental addition of data samples from each distribution. The \textcolor{orange}{\textbf{orange}} boxes indicate the data distribution currently being added for adaptation. The \textcolor{black}{\textbf{black}} boxes represent data distributions that have already been incorporated and are under fine-tuning, while the \textcolor{gray}{\textbf{gray}} boxes denote unseen data distributions. The \textcolor{magenta}{magenta} lines indicate the detector trained with CE+SupCon losses, while the \textcolor{blue}{blue} lines represent the baseline trained using only the CE loss.
    }
    \label{fig:method}
\end{figure}

\newpage
\subsection{\textbf{Feature Representation:}}

\begin{figure}[!h]
    \centering
    \includegraphics[width=\linewidth]{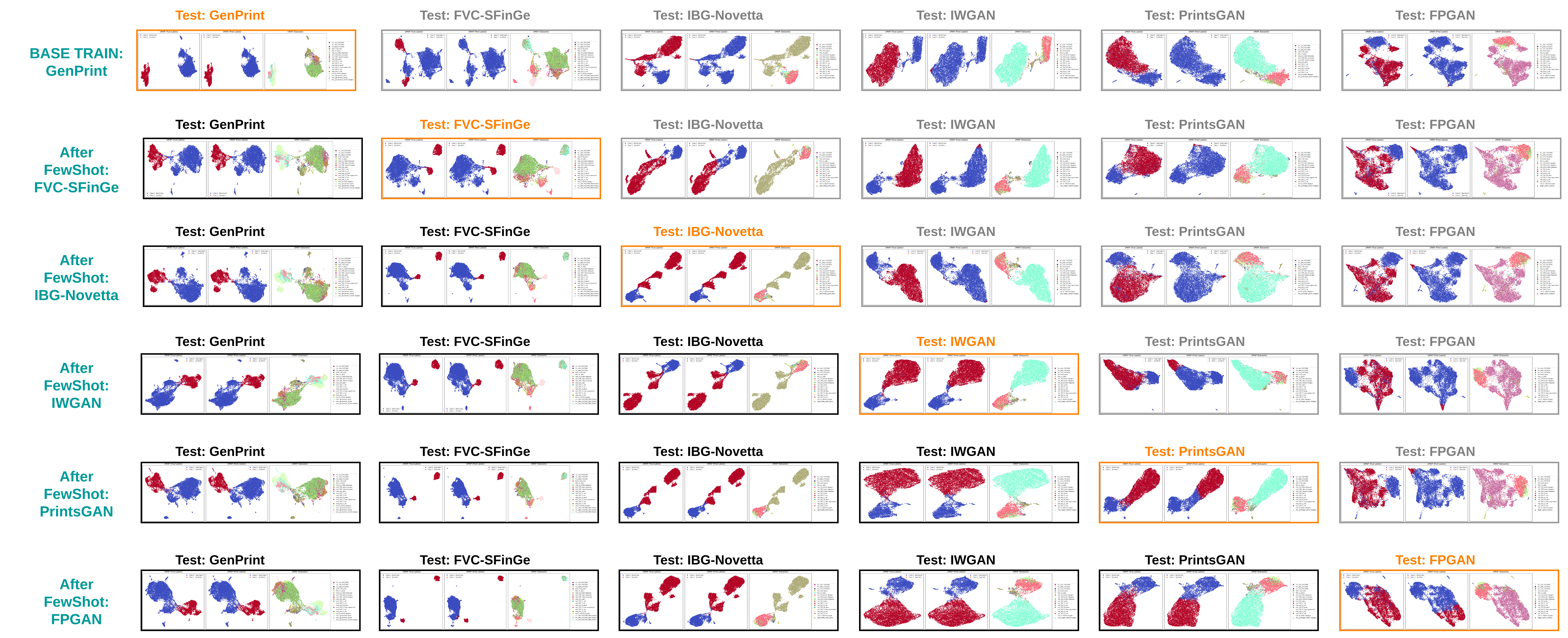}
    \caption{Plots illustrating the evolution of feature representations for synthetic vs.\ real data. Each plot shows the corresponding synthetic samples along with all real data distributions, highlighting the effect of each adaptation step. The row and column layout matches the previous ROC plot. Each plot contains three subplots: \textit{(a)} features color-coded based on ground-truth labels, \textit{(b)} features color-coded based on predicted labels, and \textit{(c)} features color-coded based on dataset name. In \textit{(a)} and \textit{(b)}, \textcolor{red}{\textbf{red}} denotes synthetic data features, while \textcolor{blue}{\textbf{blue}} denotes real data features. In \textit{(c)} colors correspond to datasets. Visualization suggests that few-shot adaptation primarily helps in forming the classifier boundary, while the underlying representation and separability remain largely unchanged and are mainly determined by the base detector.
    }
    \label{fig:method}
\end{figure}

\subsection{\textbf{Effects of Changing Adaptation Sequence:}}

\begin{figure}[!h]
    \centering
    \includegraphics[width=0.37\linewidth]{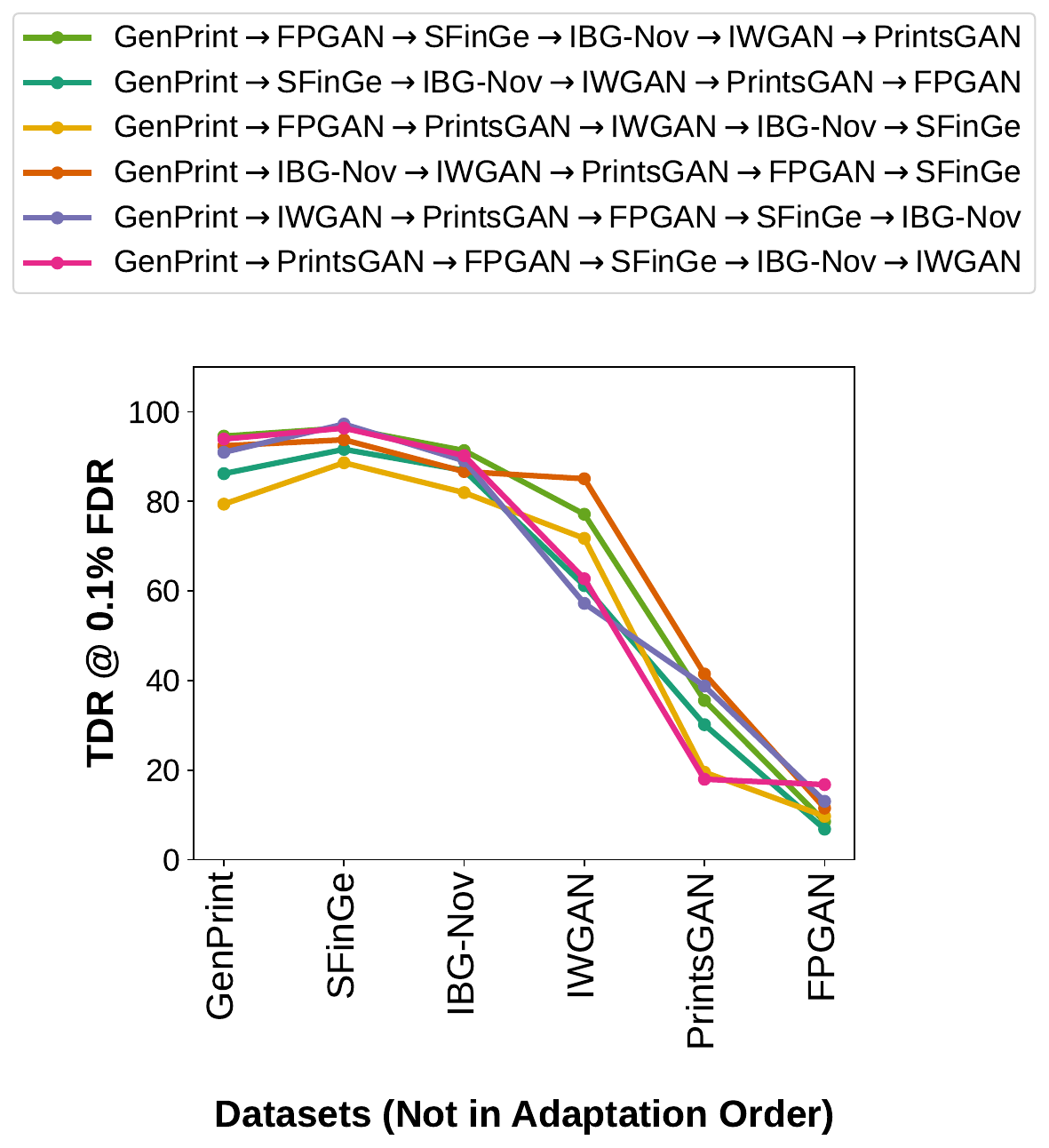}
    \includegraphics[width=0.5\linewidth]{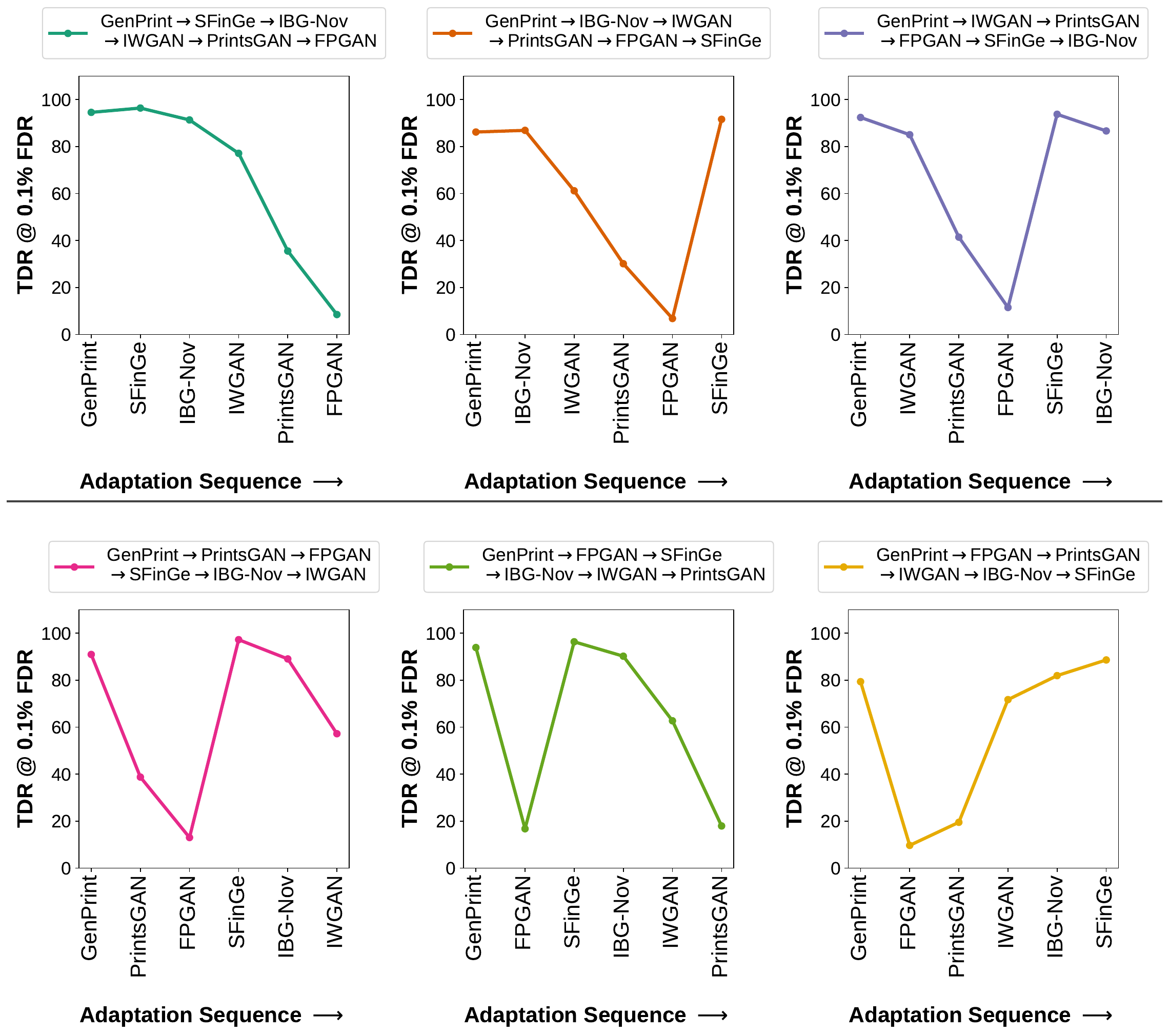}
    \caption{Figure illustrating the effect of dataset order in the adaptation sequence. Results are shown after the final adaptation step in continual training. In the left panel, each x-axis tick corresponds to a dataset, and each line represents a different adaptation sequence. The right panel organizes results with the actual adaptation sequence along the x-axis. Some variation in final performance is observed; however, it is not substantial, indicating that adaptation order has limited impact on final accuracy. In this few-shot setting, performance is primarily governed by the capacity of the base detector.
    }
    \label{fig:method}
\end{figure}

\end{document}

%% file: tables/datasets.tex
\begin{table}[!ht]
\centering
\caption{Summary of fingerprint datasets used in this study. Here, ``Finger'' and ``Image'' columns denote the number of unique fingers and the total number of images, respectively.}
\label{tab:fingerprint-datasets}
\resizebox{0.8\linewidth}{!}{%
\begin{tabular}{lcccc}
\toprule
\multirow[c]{2}{*}{\textbf{Dataset}} & \multicolumn{2}{c}{\textbf{Ink on Paper}} & \multicolumn{2}{c}{\textbf{Live Scan}} \\ \cmidrule{2-5} 
& \textbf{Finger} & \textbf{Image} & \textbf{Finger} & \textbf{Image} \\ 
\toprule
\multicolumn{2}{l}{\textbf{Real Fingerprint Datasets}} \\
\midrule
FVC-20{00},{02},{04}\cite{sfinge-cappelli2002synthetic} & - & - & 330 & 7,920 \\ 
NIST-SD4 & 2,000 & 4,000 & - & - \\ 
NIST-SD14 & 27,000 & 54,000 & - & - \\ 
NIST-SD300 & 17,659 & 17,659 & - & - \\ 
NIST-SD301 & - & - & 480 & 2,434 \\ 
NIST-SD302 & - & - & 4,000 & 13,773 \\ 
IIITD-MOLF\cite{molf-sankaran2015multisensor} & - & - & 1,000 & 12,000 \\ 
LFIW\cite{lfiw-liu2024latent} & - & - & 600 & 2,394 \\ 
 \cmidrule{1-5}
 \textbf{Total} & 46,659 & 75,659 & 6,410 & 38,521 \\ 
 
\toprule
\multicolumn{2}{l}{\textbf{Synthetic Fingerprint Datasets}} \\
\midrule
SFinGe\cite{sfinge-cappelli2002synthetic} & - & - & 330 & 2,640 \\ 
IBG-Novetta\cite{ibg_unrel-cao2018fingerprint} & - & - & 5,000 & 20,000 \\ 
IWGAN\cite{iwgan-mistry2020fingerprint} & 20,000 & 20,000 & - & - \\ 
PrintsGAN\cite{printsgan-engelsma2022printsgan} & 1,350 & 20,250 & - & - \\
FPGAN\cite{fpgan-shoshan2024fpgan} & - & - & 1,400 & 21,000 \\
GenPrint\cite{genprint-grosz2024universal} & 16,000 & 20,548 & 5,100 & 21,246 \\ 

 \bottomrule
\end{tabular}%
}
\end{table}

%% file: tables/results-table.tex
\newcommand{\repeatdot}[1]{%
  \begingroup
  \count0=0
  \loop
    $\cdot$ %
    \advance\count0 by 1
    \ifnum\count0<#1
  \repeat
  \endgroup
}
\newcommand{\diagcolor}[1]{\textcolor{orange}{#1}}
\newcommand{\grayise}[1]{\textcolor{gray}{#1}}

\begin{table*}[t]
\centering
\caption{Summary of the results from the continuous few-shot adaptation study. Here, each row represents an evolution of the detector - starting with the base detector followed by sequential adaptation to new synthetic styles. The main columns represents the performance on the test set of a known/unseen synthetic style, with the real test set remaining unchanged. The last column represents average performance over all the synthetic test sets.}
\label{tab:cross_gen_results}
\setlength{\tabcolsep}{3pt} 
\resizebox{\textwidth}{!}{%
\begin{tabular}{cl| ccc| ccc| ccc| ccc| ccc| ccc| ccc}
\toprule
\multirow{1}{*}{\rotatebox{90}{\textbf{Method}}} 
&\multirow{2}{*}{\shortstack{\ \\    \textbf{Adaptation} \\ \textbf{Sequence}}}
& \multicolumn{3}{c|}{\textbf{GenPrint}}
& \multicolumn{3}{c|}{\textbf{SFinGe}}
& \multicolumn{3}{c|}{\textbf{IBG-Novetta}}
& \multicolumn{3}{c|}{\textbf{IWGAN}}
& \multicolumn{3}{c|}{\textbf{PrintsGAN}}
& \multicolumn{3}{c|}{\textbf{FPGAN}}
& \multicolumn{3}{c}{\textbf{Mean}} \\

&& TDR & FDR & \footnotesize{TDR@}
 & TDR & FDR & \footnotesize{TDR@}
 & TDR & FDR & \footnotesize{TDR@}
 & TDR & FDR & \footnotesize{TDR@}
 & TDR & FDR & \footnotesize{TDR@}
 & TDR & FDR & \footnotesize{TDR@}
 & TDR & FDR & \footnotesize{TDR@} \\
&& $\tau$=0.5 & $\tau$=0.5 & 0.1\% FDR
 & $\tau$=0.5 & $\tau$=0.5 & 0.1\% FDR
 & $\tau$=0.5 & $\tau$=0.5 & 0.1\% FDR
 & $\tau$=0.5 & $\tau$=0.5 & 0.1\% FDR
 & $\tau$=0.5 & $\tau$=0.5 & 0.1\% FDR
 & $\tau$=0.5 & $\tau$=0.5 & 0.1\% FDR
 & $\tau$=0.5 & $\tau$=0.5 & 0.1\% FDR \\

\midrule
\multirow{6}{*}{\rotatebox{90}{ViT-CE}}
& GenPrint\scriptsize{(base)}  &  \diagcolor{99.82}  &  \diagcolor{0.65}  &  \diagcolor{96.68}  &  \grayise{0.73}  &  \grayise{0.49}  &  \grayise{0.00}  &  \grayise{2.94}  &  \grayise{0.57}  &  \grayise{0.62}  &  \grayise{1.56}  &  \grayise{0.56}  &  \grayise{0.00}  &  \grayise{0.60}  &  \grayise{0.57}  &  \grayise{0.00}  &  \grayise{0.62}  &  \grayise{0.52}  &  \grayise{0.09}               &  17.71  &  0.56  &  16.23  \\
& SFinGe                           &  99.15  &  1.14  &  88.74  &  \diagcolor{99.66}  &  \diagcolor{1.20}  &  \diagcolor{90.17}  &  \grayise{70.33}  &  \grayise{1.16}  &  \grayise{32.02}  &  \grayise{2.09}  &  \grayise{0.91}  &  \grayise{0.00}  &  \grayise{0.67}  &  \grayise{1.15}  &  \grayise{0.00}  &  \grayise{2.78}  &  \grayise{0.91}  &  \grayise{0.17}           &  45.78  &  1.08  &  35.18  \\
& IBG-Novetta                   &  98.56  &  1.66  &  82.06  &  99.91  &  1.91  &  91.54  &  \diagcolor{99.06}  &  \diagcolor{1.66}  &  \diagcolor{80.10}  &  \grayise{1.64}  &  \grayise{1.67}  &  \grayise{0.00}  &  \grayise{2.19}  &  \grayise{1.89}  &  \grayise{0.09}  &  \grayise{1.92}  &  \grayise{1.81}  &  \grayise{0.02}           &  50.55  &  1.77  &  42.30  \\
& IWGAN                         &  99.28  &  0.85  &  91.16  &  99.87  &  0.85  &  98.46  &  99.21  &  0.85  &  94.97  &  \diagcolor{94.17}  &  \diagcolor{0.89}  &  \diagcolor{44.77}  &  \grayise{20.30}  &  \grayise{0.92}  &  \grayise{3.05}  &  \grayise{2.81}  &  \grayise{1.11}  &  \grayise{0.17}        &  69.27  &  0.91  &  55.43  \\
& PrintsGAN                     &  98.99  &  1.95  &  83.71  &  100.00  &  1.94  &  96.03  &  99.29  &  2.03  &  86.86  &  97.79  &  2.09  &  32.41  &  \diagcolor{84.79}  &  \diagcolor{2.05}  &  \diagcolor{14.47}  &  \grayise{6.98}  &  \grayise{1.89}  &  \grayise{0.31}      &  81.31  &  1.99  &  52.30  \\
& FPGAN                         &  98.92  &  3.48  &  87.42  &  99.96  &  3.13  &  97.86  &  99.46  &  3.31  &  87.82  &  98.91  &  3.40  &  31.40  &  90.03  &  3.32  &  28.88  &  \diagcolor{45.86}  &  \diagcolor{3.48}  &  \diagcolor{7.47}      &  88.86  &  3.35  &  56.81  \\
\midrule
\multirow{6}{*}{\rotatebox{90}{ViT-CE+SC}}
& GenPrint\scriptsize{(base)}  &  \diagcolor{99.82}  &  \diagcolor{0.65}  &  \diagcolor{96.47}  &  \grayise{0.73}  &  \grayise{0.51}  &  \grayise{0.00}  &  \grayise{2.98}  &  \grayise{0.59}  &  \grayise{0.59}  &  \grayise{1.49}  &  \grayise{0.61}  &  \grayise{0.00}  &  \grayise{0.57}  &  \grayise{0.57}  &  \grayise{0.00}  &  \grayise{0.66}  &  \grayise{0.55}  &  \grayise{0.08}               &  17.71  &  0.58  &  16.19  \\
& SFinGe                           &  99.05  &  0.81  &  88.25  &  \diagcolor{99.66}  &  \diagcolor{0.97}  &  \diagcolor{94.49}  &  \grayise{69.83}  &  \grayise{1.03}  &  \grayise{44.64}  &  \grayise{2.02}  &  \grayise{0.77}  &  \grayise{0.00}  &  \grayise{0.64}  &  \grayise{1.03}  &  \grayise{0.00}  &  \grayise{2.25}  &  \grayise{0.79}  &  \grayise{0.16}           &  45.57  &  0.90  &  37.92  \\
& IBG-Novetta                   &  99.10  &  2.06  &  82.81  &  99.87  &  2.16  &  94.15  &  \diagcolor{99.35}  &  \diagcolor{1.99}  &  \diagcolor{81.19}  &  \grayise{1.39}  &  \grayise{1.86}  &  \grayise{0.00}  &  \grayise{2.92}  &  \grayise{2.21}  &  \grayise{0.12}  &  \grayise{2.92}  &  \grayise{2.00}  &  \grayise{0.03}           &  50.92  &  2.05  &  43.05  \\
& IWGAN                         &  99.15  &  0.87  &  93.84  &  99.96  &  0.94  &  97.99  &  99.16  &  0.85  &  94.00  &  \diagcolor{94.52}  &  \diagcolor{0.91}  &  \diagcolor{71.48}  &  \grayise{20.18}  &  \grayise{0.98}  &  \grayise{3.36}  &  \grayise{2.81}  &  \grayise{1.08}  &  \grayise{0.19}        &  69.30  &  0.94  &  60.14  \\
& PrintsGAN                     &  98.66  &  1.84  &  85.82  &  99.91  &  1.77  &  95.68  &  99.22  &  1.87  &  88.80  &  98.00  &  1.80  &  53.06  &  \diagcolor{85.96}  &  \diagcolor{1.77}  &  \diagcolor{23.53}  &  \grayise{6.83}  &  \grayise{1.77}  &  \grayise{0.34}       &  81.43  &  1.80  &  57.87  \\
& FPGAN                         &  98.79  &  3.79  &  87.76  &  99.91  &  3.43  &  95.68  &  99.68  &  3.47  &  86.92  &  99.30  &  3.62  &  69.48  &  92.89  &  3.71  &  37.71  &  \diagcolor{51.46}  &  \diagcolor{3.83}  &  \diagcolor{10.57}     &  90.34  &  3.64  &  64.69  \\
\bottomrule
\end{tabular}
}
\end{table*}